\documentclass[journal]{IEEEtran}
\usepackage{amsmath,amsfonts}
\usepackage{algorithmic}
\usepackage{array}
\usepackage[caption=false,font=normalsize,labelfont=sf,textfont=sf]{subfig}
\usepackage{textcomp}
\usepackage{stfloats}
\usepackage{url}
\usepackage{verbatim}
\usepackage{booktabs}
\usepackage{graphicx}
\usepackage{authblk}
\usepackage{hyperref}
\usepackage{microtype}
\hypersetup{
    colorlinks=true,
    linkcolor=blue,
    filecolor=blue,
    urlcolor=blue,
    citecolor=blue,
}
\usepackage{academicons}
\usepackage{scalerel}
\usepackage{orcidlink}
\def\BibTeX{{\rm B\kern-.05em{\sc i\kern-.025em b}\kern-.08em
    T\kern-.1667em\lower.7ex\hbox{E}\kern-.125emX}}
\usepackage{balance}
\begin{document}
\newcommand{\orcidicon}[1]{%
    \href{https://orcid.org/#1}{\scalerel*{\aiOrcid}{|}}}
\title{An Exploratory Deep Learning Approach for Predicting Subsequent Suicidal Acts in Chinese Psychological Support Hotlines}
\author{{Changwei Song}\orcidlink{0000-0003-3824-5663}, {Qing Zhao}\orcidlink{0000-0001-9570-9546}, {Jianqiang Li}\orcidlink{0000-0003-1995-9249}, {Yining Chen}\orcidlink{0009-0003-4319-5370}, \textsuperscript{*}{Yongsheng Tong}\orcidlink{0000-0002-1544-040X}, \textsuperscript{*}{Guanghui Fu}\orcidlink{0000-0002-6391-5983}
\thanks{Manuscript created July, 2024. This study was supported by the National Natural Science Foundation of China [82071546], Beijing Municipal High Rank Public Health Researcher Training Program [2022-2-027], the Beijing Hospitals Authority Clinical Medicine Development of Special Funding Support [ZYLX202130], and the Beijing Hospitals Authority’s Ascent Plan [DFL20221701]. The funding institutions had no role in the design, data collection, analysis, interpretation, writing of the report, or the decision to submit for publication. (Corresponding author: Yongsheng Tong and Guanghui Fu.) 

Changwei Song, Qing Zhao, Jianqiang Li, and Yining Chen are with the {School of Software Engineering, Beijing University of Technology, Beijing, China}. (e-mails: songchangwei@emails.bjut.edu.cn, zhaoqing@bjut.edu.cn, lijianqiang@bjut.edu.cn, chenyn0317@emails.bjut.edu.cn)

Yongsheng Tong is with the {Beijing Suicide Research and Prevention Center, Beijing Huilongguan Hospital, Beijing, China}, {WHO Collaborating Center for Research and Training in Suicide Prevention, Beijing, China}, and {Peking University Huilongguan Clinical Medical School, Beijing, China}. (e-mails: timystong@pku.org.cn)

Guanghui Fu is with the {Sorbonne Université, Institut du Cerveau - Paris Brain Institute - ICM, CNRS, Inria, Inserm, AP-HP, Hôpital de la Pitié Salpêtrière, Paris, France}. He is supported by a Chinese Government Scholarship provided by the China Scholarship Council (CSC).
 (e-mails: guanghui.fu@inria.fr)
}
}

%\markboth{IEEE Transactions on Affective Computing, ~Vol.~$\times$, No.~$\times$, September~$\times\times\times\times$}%
% \markboth{Submitted to IEEE Transactions on Affective Computing}%
\markboth{Preprint}%
{How to Use the IEEEtran \LaTeX \ Templates}

\maketitle

\begin{abstract}
Psychological support hotlines are an effective suicide prevention measure that typically relies on professionals using suicide risk assessment scales to predict individual risk scores.
However, the accuracy of scale-based predictive methods for suicide risk assessment can vary widely depending on the expertise of the operator. 
This limitation underscores the need for more reliable methods, prompting this research's innovative exploration of the use of artificial intelligence to improve the accuracy and efficiency of suicide risk prediction within the context of psychological support hotlines. 
The study included data from 1,549 subjects from 2015-2017 in China who contacted a psychological support hotline. Each participant was followed for 12 months to identify instances of suicidal behavior. 
We proposed a novel multi-task learning method that uses the large-scale pre-trained model Whisper for feature extraction and fits psychological scales while predicting the risk of suicide. 
The proposed method yields a 2.4\% points improvement in F1-score compared to the traditional manual approach based on the psychological scales. Our model demonstrated superior performance compared to the other eight popular models. 
To our knowledge, this study is the first to apply deep learning to long-term speech data to predict suicide risk in China, indicating grate potential for clinical applications. 
The source code is publicly available at: \url{https://github.com/songchangwei/Suicide-Risk-Prediction}.
\end{abstract}

\begin{IEEEkeywords}
Deep learning, Suicide, Mental health, Psychological support hotline, Natural language processing.
\end{IEEEkeywords}

\section{Introduction}  \label{sec:intro} 
\IEEEPARstart{T}{he} global burden of disease and the societal impact of suicide are increasingly significant, highlighting the need for effective public health interventions.
Annually, around 703,000 people worldwide die by suicide, marking it as the second leading cause of premature death among young individuals~\cite{world2021suicide}.
Despite a declining overall suicide rate in China, the prevalence of attempted suicide remains alarmingly high at approximately 0.8\%~\cite{cao2015prevalence}.
This trend is particularly concerning among adolescents, where the suicide rate continues to escalate annually~\cite{chen2018suicidal}, underscoring the critical need for effective preventive measures.
In response, psychological support hotlines, established as nationwide networks in many countries, have emerged as essential and highly effective tools in suicide prevention~\cite{gould2018follow, gould2016helping}, connecting individuals in crisis with trained professionals for immediate telephone counseling, comprehensive risk assessment, and necessary interventions~\cite{ witte2010assessing}.
However, employing suicide risk assessment scales for predicting suicide risk has shown certain limitations. 
Firstly, the positive predictive value of these scales in prospective studies tends to be moderate~\cite{quinlivan2017predictive}.
Although sensitivity and specificity metrics may appear relatively ideal, the infrequency of suicide attempts signifies that most individuals identified as high-risk do not commit suicidal behavior, with an even smaller proportion among those labeled as low-risk.
Secondly, the outcome quality of these assessments is heavily influenced by the operator's experience and capabilities.
Moreover, establishing a trusting relationship with the operator requires time for individuals seeking help for psychological issues.
This process may lead to intentional concealment or non-serious responses, adversely affecting the authenticity of the assessment results.
Finally, in emergency situations where structured assessment using scales is limited, the lack of critical data can complicate the measurement of job quality and pose significant challenges to accurately predicting suicide risk.

%Artificial intelligence technologies have been widely applied in suicide risk prediction.
Artificial intelligence (AI) has been widely applied in suicide detection.
Nobles et al.~\cite{nobles2018identification} constructed a CNN based model to detect periods of suicidality in text messages from individuals with a history of suicidal thoughts and behaviors.
Du et al.~\cite{du2018extracting} innovatively applied CNNs and transfer learning to recognize suicidal ideation within a Twitter corpus.
Fu et al.~\cite{fu2021distant} proposed a distant supervision approach based on BERT~\cite{devlin-etal-2019-bert} to identify suicide risk from Chinese social media.
These studies demonstrate the promising potential of AI in suicide risk prediction. However, the methods employed are primarily text-based and cannot be directly applied to speech data, for instance psychological support hotlines which is related to suicide behaviour.
Scherer et al.~\cite{scherer2013investigating} developed a machine learning model to predict future suicides among adolescents, utilizing interview recordings from 16 teenagers, with suicide defined as any occurrence within the six months prior to the interview. 
Amiriparian et al.~\cite{amiriparian2024enhancing} established a suicide detection model based on audio data from 20 patients responding to specific content, with suicide determined by clinician assessments. Belouali et al.~\cite{belouali2021acoustic} created an AI model to predict suicides among U.S. military personnel, using recordings from 124 veterans, with suicide defined by responses to whether they had experienced suicidal thoughts or behaviors in the past two weeks.
These studies indicate that speech data play an important role in predicting suicide risk.

However, these studies have limitations. 
Firstly, the data in these studies were not sourced from clinical settings but from interviews~\cite{scherer2013investigating, belouali2021acoustic} or speech recordings from scales~\cite{amiriparian2024enhancing}, and they all had small sample sizes (16 teenagers in~\cite{scherer2013investigating}, 20 patients in~\cite{amiriparian2024enhancing}, 124 veterans in~\cite{belouali2021acoustic}) compared to our 1,549 subjects. More importantly, our research's prediction target (suicidal behavior) was confirmed by follow-up, whereas others had not been validated by real-life outcomes but were based on previous suicidal behavior~\cite{scherer2013investigating, belouali2021acoustic} or psychological scales~\cite{amiriparian2024enhancing}. 
Also, the description of data features is often vague, such as the absence of details regarding the duration of voice recordings, which may hinder reproducibility.

% Lastly, the experimental samples are biased towards specific populations, such as adolescents or military personnel.

In this study, we conducted analysis using data from China's largest psychological support hotline, Beijing Suicide Research and Prevention Center, Beijing Huilongguan Hospital. To the best of our knowledge, we utilized the largest clinical dataset, consisting of speech recordings from 1,549 subjects, to perform the suicide prediction task. 
The speech we processed averaged around one hour, presenting a significant technical challenge. To address this, we developed a deep learning model based on the foundation model. In contrast, the machine learning techniques used in other studies~\cite{scherer2013investigating, belouali2021acoustic, amiriparian2024enhancing} are more dated and may not be up to the challenge.
We experimented with nine advanced deep learning models, and the proposed method achieved an F1-score of 71.15 [71.01, 71.34] (in \%, mean with 95\% bootstrap confidence interval), which is a 2.4\% improvement compared to traditional manual methods in clinical settings.

\section{Related work}  \label{sec:related} 
\subsection{Speech emotion analysis} \label{sec:related:speech_emotion}
The speech emotion analysis methods can be categorized into two main approaches: those leveraging handcrafted features and those employing spectrogram-based methods.

For handcrafted feature-based approaches, the process generally involves two steps. Initially, a set of speech features such as MFCC~\cite{davis1980comparison}, LPCC~\cite{atal1974effectiveness}, LFPC~\cite{itakura1968analysis}, eGeMAPS~\cite{eyben2015geneva}, ComParE~\cite{schuller2013interspeech}, and IS09~\cite{schuller2003hidden} is extracted. 
Following this, various methods for speech emotion recognition are explored. Traditionally, techniques like Gaussian Mixture Models (GMM)~\cite{mclachlan2019finite}, K-nearest Neighbors (KNN)~\cite{cover1967nearest}, and Support Vector Machines (SVM)~\cite{burges1998tutorial} were popular. 
However, the advent of deep learning has led to the integration of methods based on this technology. 
For instance, Kumbhar et al.~\cite{kumbhar2019speech} utilized LSTM networks to analyze MFCC features from speech frames, learning the contextual information within the speech for emotion recognition. 
Similarly, Li et al.~\cite{li2021speech} introduced the BiLSTM-DSA model, which applies attention mechanisms to both forward and backward output vectors of a BiLSTM network. 
Emotion recognition is then performed by analyzing these feature embedding. 
Other notable methods include the EmNet by Kim et al.~\cite{kim2018emotion}, which combines CNN and RNN models, and the Attention-BLSTM-FCN model by Zhao et al.~\cite{zhao2018exploring}, illustrating the diversity of approaches in handcrafted feature-based emotion recognition.

The success of CNN in image classification has inspired researchers to explore their application in speech emotion recognition using spectrograms. 
This method typically involves converting speech signals into spectrograms, followed by feature extraction using CNNs. 
The objective is to leverage CNNs' ability to identify relevant features for emotion recognition. 
Early research by Cummins et al.~\cite{cummins2017image} demonstrated this approach using AlexNet and VGG19 networks. 
Li et al.~\cite{li2018attention} proposed a novel dual-branch CNN to extract both frequency and time domain information from spectrograms, further refining the emotion recognition process. 
Neumann et al.~\cite{neumann2019improving} combined CNN-extracted features with those obtained through unsupervised learning from unlabeled data, showcasing the potential of CNNs in enhancing speech emotion recognition.

Despite these advancements, most existing models focus on recognizing emotions from individual sentences, without considering the context of binary dialogues. 
Limited research, such as the studies by Liscombe et al.~\cite{liscombe2005using} and Wöllmer et al.~\cite{wollmer2010context}, has shown that incorporating contextual information can significantly improve emotion recognition accuracy. 
Zhang et al.~\cite{zhang2017interaction} further explored this by proposing an emotion interaction transition model, highlighting the importance of context in emotion recognition.

However, these models are not suited for the task addressed in this paper, as they are designed for speech segments shorter than 30 seconds, while our data comprises telephone recordings lasting from 9 minutes to 160 minutes. 
The challenges lie in the need for a vast number of model parameters to handle such lengthy data and the absence of a temporal model capable of capturing context dependencies over extended periods.

\subsection{AI-based suicide risk prediction} \label{sec:related:ai_suicide}
Current research on automated suicide detection primarily utilizes artificial intelligence (AI) technologies, including machine learning (ML) and deep learning (DL), to assess suicide risk levels through the analysis of structured, speech, and textual data. 
AI methods can identify patterns and indicators related to suicide risk, enabling more accurate and timely automated predictions. 
Structured data typically includes clinical records and demographic information, while speech and textual data are derived from patient interviews, social media posts, and other communication channels.
 
Suicide detection based on structured data typically involves two steps~\cite{barak2017predicting,nock2022prediction,lyu2022prediction}: first, the collection of structured data on risk factors such as demographic characteristics, prescription medications, mental symptoms, history of suicidal thoughts and behaviors, family history of mental disorders, and traumatic life events; second, employing machine learning methods for detection suicide risk, including logistic regression~\cite{cox1958regression}, decision trees~\cite{loh2011classification}, Naive Bayes classifier~\cite{john2013estimating}, and multi-layer perceptrons (MLP)~\cite{rosenblatt1961principles}.
However, structured data  poses challenges due to its limited availability and insufficient information.
Suicide detection based on text data often utilizes online media data.
For example, Guan et al.~\cite{guan2015identifying} used simple logistic regression (SLR)~\cite{cox1958regression} and random forest (RF)~\cite{breiman2001random} algorithms to identify high-suicide-risk individuals among Chinese Weibo users. 
And Tadesse et al.~\cite{tadesse2019detection} utilized an LSTM-CNN model with word embeddings for suicide prediction on Reddit. 
These studies indicate that AI technology shows promising potential in suicide risk prediction. 
However, the existing methods predominantly rely on textual data, which are not directly applicable to the speech data utilized in Chinese psychological support hotlines.

Some recent studies have integrated computer algorithms for speech analysis in suicide risk prediction.
Scherer et al.~\cite{scherer2013investigating} developed machine learning models for suicide classification in adolescents. They collected an interview corpus from 16 teenagers aged 13 to 17, with interview durations averaging around 778 seconds for suicidal adolescents and 451 seconds for non-suicidal adolescents. Using the developed Hidden Markov Model (HMM)~\cite{rabiner1986introduction}, they achieved an accuracy of 81.25\% at the interview level and 69\% at the utterance level.
Amiriparian et al.~\cite{amiriparian2024enhancing} developed machine learning method (SVM) using speech data and meta-features for suicide ideation detection. They collected data from 20 clinical subjects and recorded three types of speech: word repetition, picture description, and pronunciation, with audio lengths of less than 30 seconds. Additionally, meta-information and mental health scales were collected. Suicide ideation was identified by a clinical doctor using a Likert scale ranging from 1 to 6 (with 1-4 indicating low risk and 5-6 indicating high risk). The results indicated that the model built using deep features achieved a balanced accuracy of 66.2\%, while the performance improved to 94.4\% when combined with meta-information and mental health scales.
Belouali et al.~\cite{belouali2021acoustic} developed machine learning models for suicide ideation prediction in U.S. veterans. They collected 588 audio recordings from 124 subjects via a mobile app, with each recording averaging around 44 seconds in length. Suicide ideation was defined using self-report psychiatric scales and questionnaires. They built several machine learning algorithms, incorporating both language and acoustic features as inputs. Their models achieved high performance in the suicide ideation task, with an AUC of 80\%.

In our study, we used the largest dataset derived from a clinically utilized psychological support hotline, and we confirmed the gold standard through follow-up.
These three highly related studies~\cite{scherer2013investigating,amiriparian2024enhancing,belouali2021acoustic} involved significantly fewer subjects compared to our research.
From the perspective of gold standard determination, we confirmed whether the subjects exhibited suicidal behavior through 12-month follow-ups. In contrast, other studies determined this based on previous suicidal behavior~\cite{scherer2013investigating,belouali2021acoustic}, psychological scales~\cite{amiriparian2024enhancing}, or experienced suicidal thoughts~\cite{belouali2021acoustic}. Although the purposes of these studies varied, our criteria were confirmed by real-life validation rather than uncertainties such as suicidal thoughts or scales, giving us an advantage over other studies.
Also, our dataset comprises long-duration audio recordings, with an average length of 59.6 minutes, presenting unique challenges for our model design. In contrast, the audio lengths in other studies are much shorter— a maximum of 778 seconds in Scherer et al.\cite{scherer2013investigating}, around 30 seconds in Amiriparian et al.\cite{amiriparian2024enhancing}, and around 44 seconds in Belouali et al.~\cite{belouali2021acoustic}. 
Therefore, we employ the latest methods to address these unique challenges, as the traditional methods used in~\cite{scherer2013investigating, amiriparian2024enhancing, belouali2021acoustic}, such as Hidden Markov Models and SVMs, are not applicable to our task.
Moreover, our experimental sample is more representative of the general population. 
In contrast, Belouali et al.~\cite{belouali2021acoustic} focused on veterans, while Scherer et al.~\cite{scherer2013investigating} focused on adolescents.

\section{Workflow of the Beijing psychological support hotline} \label{sec:background}
\begin{figure*}[!hbtp]
\centering
\includegraphics[scale=0.75]{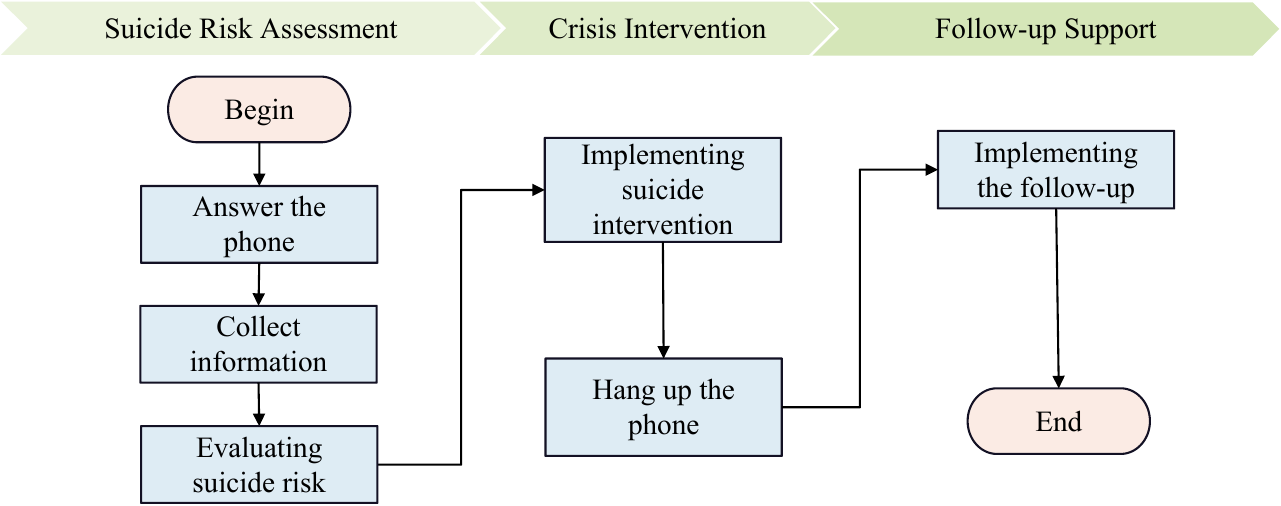}
\caption{The flowchart of the psychological support hotlines. The service process of a psychological support hotlines is divided into three main stages: suicide risk assessment, crisis intervention, and follow-up support. During the suicide assessment stage, interventionists usually spend 15 to 30 minutes using a suicide risk assessment scale to determine whether the caller has suicidal tendencies. In the crisis intervention stage, for those callers assessed as having suicidal tendencies, necessary support and assistance will be provided. Finally, in the follow-up support stage, these individuals will undergo continuous risk assessment and intervention over the next 12 months to ensure that they receive long-term and sustained care and support.}
\label{fig:Hotline}
\end{figure*}

The Beijing psychological support hotline platform is the first psychological support hotlines in China, based at the Beijing Suicide Research and Prevention Center in Beijing Huilongguan Hospital. 
The operation of psychological support hotlines can be divided into three sequential stages: suicide risk assessment, crisis intervention, and follow-up support.
For simplicity, the staff operating the psychological intervention hotline will be referred to as ``Operator'', and individuals calling the hotline will be referred to as ``Caller''. 
All \textgreater 30 hotline operators were trained in assessing suicidal risk before answering calls independently.
The following sections offer a detailed description of the psychological intervention process. 
The corresponding flowchart is illustrated in Figure~\ref{fig:Hotline}, and Tong et al.\cite{tong2023predictive} details the full process.

\subsection{Suicide risk assessment and crisis intervention} \label{sec:background:hotline}

The initial suicide risk assessment phase involves building basic trust between the operator and caller to gather essential information about their background, current circumstances, psychological state, and any suicidal thoughts or behaviors.
Following this, suicide risk assessment scales developed by Tong et al.~\cite{tong2020prospective, tong2023predictive} are employed to objectively evaluate the caller's suicide risk scores.
These scales use standardized questions and criteria to identify individuals at increased risk by examining factors such as past and present suicidal behavior, symptoms of depression, and feelings of hopelessness, among others.
This scale features 31 items that evaluate 12 critical elements of suicide risk.
The scale assesses suicide risk using a combination of continuous variables, such as suicidal ideation and plans, acute life events, and ten dichotomous variables elements, generating an aggregate score from 0 to 16. Scores from 0 to 7 indicate low-moderate suicidal risk, while 8 to 16 suggest high risk~\cite{tong2020prospective, tong2023predictive,belouali2021acoustic}. Incomplete data from respondents failing to answer more than five items results in missing scores. 
Typically, it takes 10-15 minutes to complete the scale.
Table~\ref{tab:scale} lists these elements of the scale; for further details, see references~\cite{tong2020prospective, tong2023predictive}.

Subsequently, the crisis intervention phase focuses on immediate risk mitigation and support. 
Note that ideally, the scale is administered first, followed by crisis intervention. However, if the caller is exhibiting high suicidal ideation, the operator must either provide immediate crisis intervention or be flexible in the sequence, incorporating scale questions as needed.

\begin{table}[!ht]
\begin{center}
\caption{Suicide risk assessment scale used in Beijing psychological support hotline, developed by Tong et al.~\cite{tong2020prospective, tong2023predictive}.}
\label{tab:scale}
\begin{tabular}{l c}
\hline
\textbf{Elements (number of items) } & \textbf{Score} \\
\hline
Suicidal ideation and plan (3) & 0/1/4 \\
Severe depression (11) & 0/1 \\
Hopelessness (1) & 0/1 \\
Psychological distress (1) & 0/1 \\
Acute life events (2) & 0/2 \\
Chronic life events (2) & 0/1 \\
Alcohol or substance misuse (3) & 0/1 \\
Severe physical illness (1) & 0/1 \\
Fear of being attacked (2) & 0/1 \\
History of being abused (2) & 0/1 \\
Suicide attempt history (1) & 0/1 \\
Relatives or acquaintances suicidal acts history (2) & 0/1 \\
\hline
\end{tabular}
\end{center}
\end{table}

\subsection{Twelve months follow-up support}
The final phase, follow-up support, ensures the caller receives continued assistance and monitoring to prevent any resurgence of suicide risk.
All callers designated for follow-up at 6-and 12-months after the initial assessment. And in specify, depending on the level of crisis, the operator schedules follow-up visits with varying frequencies. 
We defined suicide as the act of deliberately killing oneself with the intent to die, and a suicide attempt as the intentional act of non-fatal self-harm with the intent to die. Instances of self-harm without any suicidal intention were excluded.
Note that the final label that our model learned is the presence or absence of suicidal behavior in the subject during the 12 months follow-up. 
 
\section{Methods} \label{sec:methods}
\begin{figure*}[!hbtp]
\centering
\includegraphics[scale=0.75]{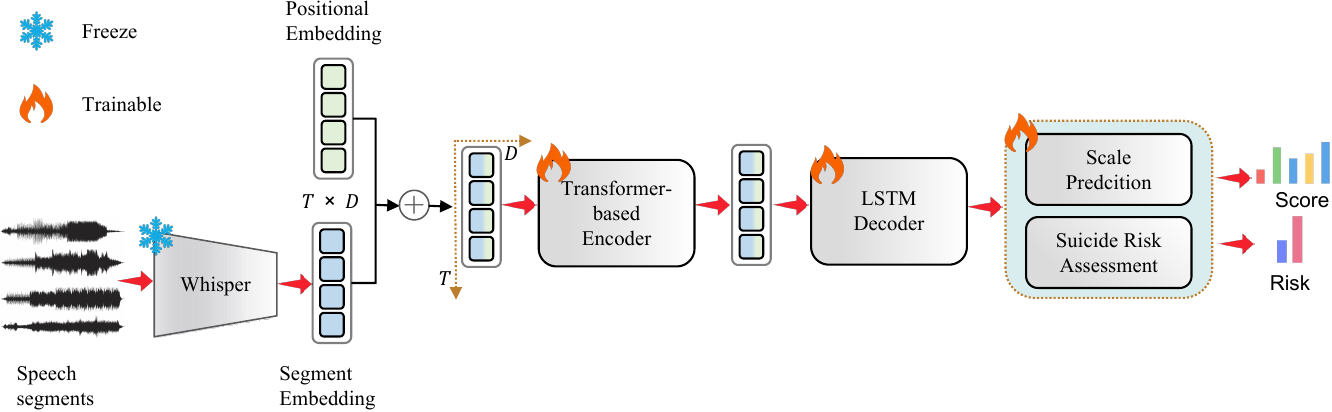}
\caption{The model architecture for this research. First, segmented speech is extracted as a series of embeddings using the pre-trained Whisper model. These feature embeddings are then processed through a transformer-based encoder to establish long-distance dependencies. Finally, the LSTM decoder performs multi-task learning by connecting to two classification heads—one for scale prediction and the other for overall suicide risk assessment.}
\label{fig:model_architecture}
\end{figure*}

The proposed model is divided into three parts: initially, the segmented long speech sequences are extracted as embeddings by the pre-trained Whisper~\cite{whisper_radford2023robust} model. Subsequently, a transformer~\cite{transformer_vaswani2017attention} based encoder learns the dependencies among these long sequence embeddings. Finally, an LSTM~\cite{lstm_hochreiter1997long} based decoder is trained in a multi-task learning way, aiming to fit both the suicide risk assessment scale and the overall suicide risk. The model architecture is illustrated in Figure~\ref{fig:model_architecture}.

\subsection{Whisper based long sequence speech feature extractor} \label{sec:methods:feature_extractor}
Whisper~\cite{whisper_radford2023robust} is a large-scale pre-trained model designed for speech processing, trained using 680,000 hours of multilingual and multitask supervision. Whisper was developed to predict the raw text of transcripts, leveraging the capabilities of sequence-to-sequence models to map between spoken utterances and their textual representations. The architecture of Whisper employs an end-to-end encoder-decoder Transformer architecture. It demonstrates excellent zero-shot performance across a variety of diverse datasets, with capabilities in multilingual speech recognition, speech translation, and language identification. Additionally, it can serve as a backbone for feature extraction.
For our study, we utilized Whisper as a feature extractor by removing its task-specific heads, using the output from the final layer as the speech embeddings for our analysis.

In our research, we initially divide a speech from the psychological support hotlines $S$ into $n$ segments as $S = [s_1, s_2, ..., s_n]$. In our experiments, we divide it into 30-second chunks. Whisper acts as the feature extractor $W$ to generate each segment as a set of feature embeddings $E$, where the dimension of the embedding is $d$, as shown in Equation~\ref{eq:feature_extract}.
\begin{equation}
\label{eq:feature_extract}
\begin{aligned}
    E^d &= W(S) \\
    E^d &= [e_1, e_2, ..., e_n]
\end{aligned}
\end{equation}

\subsection{Transformer based encoder} \label{sec:methods:encoder}
The encoder aims to extract global features and reveal intricate relationships within sequences, establishing the semantic high-level feature for the decoder to generate output.
The encoder section follows the architecture of the Transformer~\cite{transformer_vaswani2017attention}. We incorporate absolute positional encoding to enable the model to learn the relative positional connections among speech segments in the sequence.
The model takes the feature embeddings $E^d$ as input and combines them with their respective positional information. 
Absolute positional encoding employs sine and cosine functions of varying frequencies to encode distinct positions. 
Let $t$ represent the desired position in a speech, where $\vec{p_t} \in E^d$ is its corresponding encoding. The function $f$ that produce the positional embedding represent as Equation~\ref{eq:position_embed}.
\begin{equation}
\label{eq:position_embed}
\begin{aligned}
    \vec{p_t}^{(i)} = f(t)^{(i)} & := 
  \begin{cases}
      \sin({\omega_k} . t),  & \text{if}\  i = 2k \\
      \cos({\omega_k} . t),  & \text{if}\  i = 2k + 1
  \end{cases}
\end{aligned}
\end{equation}
where 
\begin{equation}
\omega_k = \frac{1}{10000^{2k / d}}
\end{equation}
Finally the feature embedding after merged with positional embedding ${e_t}'$ represented as Equation~\ref{eq:merge_embed}.
\begin{equation}
\label{eq:merge_embed}
{e_t}'= e_t + \vec{p_t}
\end{equation}

The merged feature embedding ${e_t}'$ is then passed through the self-attention module to establish the relationship between inputs and outputs.
% The positionally encoded input $\mathbf{x'}_{\text{pos}}$ is then fed into the self-attention mechanism. 
Within each head of the self-attention layer, the computation is as follows:
\begin{equation}
\text{Attention}(Q, K, V) = \text{softmax}\left(\frac{QK^T}{\sqrt{d_k}}\right)V,
\label{eq:self_attention}
\end{equation}
where $Q$, $K$, and $V$ represent the Query, Key, and Value matrices, respectively. 
They are obtained by multiplying the positionally encoded input $[e^{'}_1, e^{'}_2, ..., e^{'}_n]$ with different weight matrices $W^Q$, $W^K$, $W^V$. $d_k$ denotes the dimension of the key vectors. 
This attention mechanism allows the model to focus on different parts of the input sequence such that each element can dynamically adjust its representation in the context of the entire sequence during processing.
In practice, to enhance the model's learning capability further, a multi-head attention mechanism is often employed. 
This approach calculates multiple distinct attention heads in parallel, then concatenates their outputs, as follows:
\begin{equation}
\text{MultiHead}(Q, K, V) = \text{Concat}(\text{head}_1, \ldots, \text{head}_h)W^O,
\label{eq:multihead}
\end{equation}
where each head $(\text{head}_i)$ is an independent attention function:
\begin{equation}
\text{head}_i = \text{Attention}(QW^Q_i, KW^K_i, VW^V_i)
\label{eq:head}
\end{equation}
Here, $W^Q_i$, $W^K_i$, and $W^V_i$ are the Query, Key, and Value weight matrices corresponding to the $i^{th}$ head, respectively. 
$W^O$ is the weight matrix used for output projection after concatenating the attention outputs of all heads.
The output of the multi-head attention mechanism is subsequently passed to a Feed Forward Network (FFN), which is identical for each position and processes independently:
\begin{equation}
\text{FFN}(\hat{\mathbf{x}}') = \max(0, \hat{\mathbf{x}}'W_1 + b_1)W_2 + b_2,
\label{eq:ffn}
\end{equation}
where $W_1$ and $W_2$ denote the weight parameters of the feed forward network, while $b_1$ and $b_2$ are the corresponding bias terms.
The Transformer Encoder employs Layer Normalization and Residual Connection to speed up training and enhance the model's capacity for deep representations. 
The output of each part of the self-attention and feed forward network takes the form:
\begin{equation}
\text{LayerNorm}(\mathbf{x} + \text{Sublayer}(\mathbf{x}))
\label{eq:layernorm}
\end{equation}
where $\text{Sublayer}(\mathbf{x})$ represents the output from either the self-attention or the feed forward network.
The final output of the module is denoted as $\tilde{\mathbf{x}}  \in \mathbb{R}^{L \times D}$.

\subsection{LSTM based multi-task decoder} \label{sec:methods:decoder}
The decoder consists of an LSTM network followed by two task-specific heads: a ``Score Head'' for predicting the scores on a suicide risk assessment scale and a ``Risk Head'' for predicting suicide risk levels. 
The LSTM in the decoder plays a crucial role in receiving outputs from the encoder and utilizing them to integrate the global context of the input sequence. 
This process ensures that the generated sequence is not merely a cumulative result of single-step predictions but a coherent output reflecting the comprehensive meaning of the entire input sequence.
In the LSTM networks, the update process involves three main gates: the forget gate, the input gate, and the output gate, as well as a cell state. 
Given a sequence length $L$, a hidden state dimension $D$, for an input at time step $t$, $\tilde{\mathbf{x}}_t \in \mathbb{R}^D$, the previous time step's hidden state $h_{t-1} \in \mathbb{R}^D$, and cell state $C_{t-1} \in \mathbb{R}^D$, the update of an LSTM unit can be described by the following equations:
Forget gate (decides what information is discarded):
\begin{equation}
f_t = \sigma(W_f \cdot [h_{t-1}, \tilde{\mathbf{x}}_t] + b_f)
\label{eq:forget_gate}
\end{equation}
Input gate (decides what information to update into the cell state):
\begin{equation}
i_t = \sigma(W_i \cdot [h_{t-1}, \tilde{\mathbf{x}}_t] + b_i)
\label{eq:input_gate}
\end{equation}
Candidate cell state (prepares for the update of the cell state):
\begin{equation}
\tilde{C}_t = \tanh(W_C \cdot [h_{t-1}, \tilde{\mathbf{x}}_t] + b_C)
\label{eq:candidate_cell_state}
\end{equation}
Update cell state (combines information from the forget gate and the input gate to update the cell state):
\begin{equation}
C_t = f_t * C_{t-1} + i_t * \tilde{C}_t
\label{eq:update_cell_state}
\end{equation}
Output gate (decides the next hidden state):
\begin{equation}
o_t = \sigma(W_o \cdot [h_{t-1}, \tilde{\mathbf{x}}_t] + b_o)
\label{eq:output_gate}
\end{equation}
Update hidden state:
\begin{equation}
h_t = o_t * \tanh(C_t)
\label{eq:update_hidden_state}
\end{equation}
In these equations, $\sigma$ denotes the sigmoid activation function, which limits the output of gates between 0 and 1; $\tanh$ represents the hyperbolic tangent activation function, outputting values between -1 and 1; $\cdot$ signifies matrix multiplication; (*) denotes the Hadamard product (element-wise multiplication); $W$ and $b$ are model parameters representing weight matrices and bias terms, respectively.
The final output of the LSTM is denoted as $\mathbf{h}' \in \mathbb{R}^{L \times D}$.

Subsequently, a Score Head and a Risk Head are employed to predict the suicide risk assessment scale score and the level of suicide risk, respectively, based on $\mathbf{h}'$.
First, $\mathbf{h}'$ is aggregated into a single vector using an average pooling layer, as follows:
\begin{equation}
\boldsymbol{z} = \frac{1}{L}\sum_{i=1}^L \boldsymbol{h}_i
\label{eq:average_pooling}
\end{equation}
For the scale prediction task, we employ a fully connected layer and the softmax function, as follows:
\begin{equation}
y_s = \text{Softmax}(W_s \boldsymbol{x} + b_s)
\label{eq:suicide_risk_assessment}
\end{equation}
For the suicide risk assessment task, we use a fully connected layer and the Sigmoid function to obtain probabilities for high risk and low risk, with the formula:
\begin{equation}
y_r = \text{Sigmoid}(W_r \boldsymbol{z} + b_r)
\label{eq:suicide_risk_prediction}
\end{equation}

\subsection{Loss function} \label{sec:methods:Loss_Function}
The loss function is a hybrid of binary cross-entropy and categorical cross-entropy loss functions, as demonstrated in Equation~\ref{eq:multi_task_loss}. 
In this equation, the parameters $\alpha$ and $\beta$ represent the weights of the two loss functions in the total loss, respectively. 
This weighting approach allows the model to balance the importance of binary classification tasks and multi-class classification tasks during the training process.
\begin{equation}
\begin{aligned}
L_{r}(y_r, \hat{y_r}) &= -\left(y_r \log(\hat{y_r}) + (1 - y_r) \log(1 - \hat{y_r})\right), \\
L_{s}(y_s, \hat{y_s}) &= -\sum_{s=1}^{M} y_s \log(\hat{y}_s), \\
L &= \alpha L_{r} + \beta L_{s},
\end{aligned}
\label{eq:multi_task_loss}
\end{equation}
where $y_r$ represents the actual suicide risk, $\hat{y}_r$ denotes the predicted suicide risk, $y_s$ indicates the actual suicide risk assessment scale score, and $\hat{y}_s$ signifies the predicted suicide risk assessment scale score.

\section{Experiments} \label{sec:experiments}

\subsection{Dataset}
\begin{figure}[!hbtp]
\centering
\includegraphics[width=1.0\linewidth]{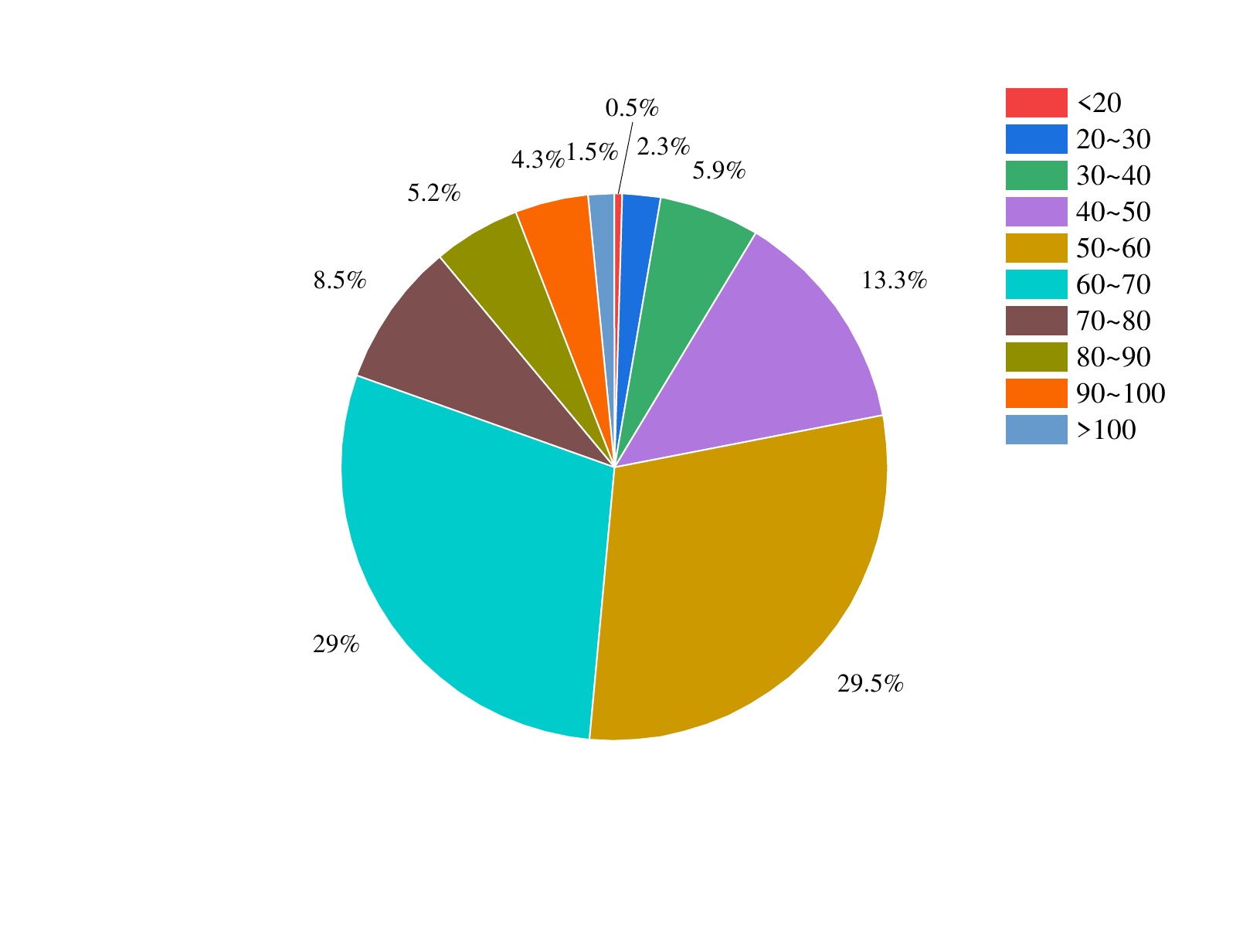}
\caption{Distribution of speech duration (in minutes) for the psychological support hotlines.}
\label{fig:dataset}
\end{figure}

% The study was approved by the Institutional Ethical Committee (Piti\'e Salp\^etri\`ere Hospital, Ile-de-France VI, n\textdegree DC-2009-957) and by the French Data Protection Authority (CNIL, Commission Nationale de l’Informatique et des Libert\'es, DR 2013-279).
%(657 males and 892 females)
All the data were taken from the psychological support hotlines of Beijing Suicide Research and Prevention Center, Beijing Huilongguan Hospital from 2015-2017.  
In our research, 8,838 callers were enrolled retrospectively with a 12-month follow-up, and 746 of them (8.4\%) had a history of suicidal behavior confirmed during the follow-up. To maintain a balanced dataset for model development, we randomly selected 803 subjects who did not exhibit suicidal behavior during the follow-up.
We included a total of 1,549 subjects (42.4\% males and 57.6\% females), aged 11-67 years (mean 26 years), with most subjects between the ages of 17 and 30 years. 
% We selected the speech for complete follow-up interviews. 
% Cases of suicide were confirmed through follow-up interviews.
% Among them, 746 subjects had committed suicidal acts within a 12-month period, while the remaining 803 did not.
The durations of these speech range from 9 minutes to 120 minutes, with a total duration of approximately 1538.6 hours, averaging around 59.6 minutes. 
The distribution of speech can be seen in Figure~\ref{fig:dataset}.
The dataset was splitted into training and test sets at a ratio of 4:1. Further, within the training set, a secondary split was conducted to a subset for validation, also at a 4:1 ratio. 
The data distribution can be seen in Table~\ref{tab:dataset}.

As introduced in Section~\ref{sec:background}, completing the suicide risk assessment process typically requires 10-15 minutes, while the overall communication process extends from 15 to 30 minutes, immediately followed by the implementation of relevant crisis interventions. Based on this, the first 30 minutes of a call was identified as a critical timeframe for predicting suicide risk. Consequently, in our experiments, we captured the initial 30-minute segment of calls involving prolonged speech and used zero-padding to standardize data for calls that lasted less than 30 minutes.

\begin{table}[!ht]
\centering
\caption{Distribution of the datasets}
\begin{tabular}{lcc}
\hline
Dataset & Suicide & No suicide \\ \hline
Training set   & 476 & 516 \\
Validation set & 115  & 132  \\
Test set       & 155 & 155 \\ \hline
\end{tabular}
\label{tab:dataset}%
\end{table}

\subsection{Comparison methods} \label{sec:experiments:Baseline}
We compare the traditional manual scale rating approach with a variety of popular deep learning models, which are described in detail as follow:
\begin{itemize}
    \item \textbf{Manual scale rating:} As introduced in Section~\ref{sec:background}, operators assess callers using suicide risk assessment scales, categorizing them into high and low suicide risk groups. To evaluate the accuracy of the deep learning model relative to the traditional manual approach, we compared the manual scale ratings with the model's predictions on the test set.
    
    \item \textbf{LSTM~\cite{lstm_hochreiter1997long}:} Long Short-Term Memory (LSTM) networks, a Recurrent Neural Networks (RNNs) variant, address the long-term dependency issue in RNNs. LSTMs manage information flow using three gates: forget, input, and output, facilitating the retention or disposal of information based on its relevance to the current context and predictions.

    \item \textbf{BiLSTM~\cite{graves2005framewise}:} Bidirectional LSTM (BiLSTM) processes data both forwards and backwards, employing two separate LSTMs. This dual-directional data processing enables the model to utilize past and future context, enhancing its performance in tasks requiring comprehensive context understanding.

    \item \textbf{GRU~\cite{gru_cho2014properties}:} Gated Recurrent Unit (GRU) simplifies the LSTM structure by merging the forget and input gates into an update gate and combining the cell state with the hidden state. This reduction in complexity allows GRUs to train faster than LSTMs while maintaining similar efficacy across tasks.

    \item \textbf{BiGRU~\cite{9424566}:} Bidirectional GRU (BiGRU) merges GRU's structure with bidirectional processing, capturing contextual information from both directions. This model effectively balances complexity and performance in capturing long-range dependencies.

    \item \textbf{Transformer~\cite{transformer_vaswani2017attention}:} The Transformer models the ability to simultaneously focus on various parts of the input sequence, enhancing comprehension and processing efficiency. 

    \item \textbf{AttentionLSTM~\cite{wang2016attention}:} AttentionLSTM combines the attention mechanism with LSTM networks, enabling dynamic focusing on the most relevant input segments for the task at hand. This model excels in managing long sequences and complex dependencies by leveraging both LSTM's memory capabilities and the attention mechanism's selective focus.

    \item \textbf{AttentionGRU~\cite{chorowski2015attention}:} AttentionGRU integrates the attention mechanism with Gated Recurrent Unit (GRU) networks, enhancing the model's efficiency in sequence processing. This combination provides dynamic weight allocation and sequence handling, leading to improved predictions in sequence modeling tasks.

    \item \textbf{Mamba~\cite{mamba}:} The Mamba model employs a ``selective state space model'' for efficient and effective sequential modeling, especially over long sequences. It features a content-based selective mechanism and a hardware-aware parallel algorithm, optimizing performance through recursive processing.
\end{itemize}

\subsection{Model decision interpretability} \label{sec:experiments:interpretability}
Deep learning has a controversial decision basis due to its black-box nature, especially in the healthcare domain where security and privacy are required. Providing interpretability of the model is an important step in addressing these concerns. For this purpose, we extracted the attention weights of each speech segment from the decoder part of our model (as introduced in Section~\ref{sec:methods:encoder}). We further analyzed the top ten weighted sentences in the speech to understand the model's decision basis. As mentioned earlier, speech segmentation was performed every 30 seconds during our experiments, which may contain a lot of dialog information. Therefore, we performed English translations of the summaries without changing the original meaning in the case presentations.

\subsection{Implementation details} \label{sec:experiments:implementation}
All experiments were conducted using an NVIDIA GeForce RTX 2080 TI equipped with 12 GB of GPU memory. We employed the PyTorch~\cite{paszke2019pytorch} framework for model development. The training process consisted of 100 epochs with a batch size of 32, utilizing AdamW ~\cite{kingma2014adam} as the optimizer. 
As discussed in Section~\ref{sec:methods:feature_extractor}, the Whisper-large-v2 model served as the feature extractor, and we divide the speech recordings into 30-second chunks for feature extraction.
Each speech segment was converted into a feature embedding with 1280 dimensions. 
We did not perform any additional preprocessing to simplify our system.
For evaluation, we used precision, recall, and F1-score as metrics for our task. For each metric, we reported its mean value as well as the corresponding 95\% confidence interval (CI) computed using bootstrap over the independent test set. 
The source code of all these methods are public avaiable at:\url{https://github.com/songchangwei/Suicide-Risk-Prediction}.

\section{Results}\label{sec:results}

The results highlighted in Table~\ref{tab:result} illustrate the comparative efficacy of traditional manual scale rating, eight time-series deep learning models, and our proposed model in predicting suicide risk. 
The example cases for error analysis are presented in Table~\ref{tab:example}.

\begin{table*}[!ht]
\centering
\caption{Performance comparison of traditional manual scale rating and nine deep learning methods (include our proposed model) on suicide prediction task.}
\label{tab:model_performance}
\begin{tabular}{@{}lcccc@{}}
\toprule
Models   & Precision(\%) & Recall(\%) & F1-score(\%)  \\ \midrule
Manual scale rating & 61.42 [61.39, 61.88]    & 78.06 [78.02, 78.42] & 68.75 [68.73, 69.08]      \\
\hline
LSTM & 57.65 [57.38, 57.79]    & 82.58 [82.35, 82.72] & 67.90 [67.61, 67.95]      \\
GRU & 54.54 [54.37, 54.76]    & 85.16 [85.13, 85.50] & 66.49 [66.34, 66.68]     \\
BiLSTM & 56.88 [56.71, 57.11]   & 80.00 [79.80, 80.20] & 66.48 [66.27, 66.62]      \\
BiGRU & 54.09 [54.05, 54.46]    & 85.16 [85.04, 85.39] & 66.16 [66.07, 66.42]      \\
AttentionLSTM & \textbf{62.75 [62.34, 62.95]}    & 79.35 [79.21, 79.59] & 70.08 [69.84, 70.48]     \\
AttentionGRU & 61.01 [60.73, 61.18]    & 69.67 [69.38, 69.82] & 65.06 [64.74, 65.11]     \\
Transformer & 58.22 [58.01, 58.43]    & 84.51 [84.12, 84.89] & 68.94 [68.79, 69.23]      \\ 
Mamba & 56.11 [55.74, 56.14]    & \textbf{85.80 [85.73, 86.08]}& 67.85 [67.54, 67.87]     \\ 
\hline
Our proposed model & 61.11 [61.04, 61.45]   & 85.16 [84.91, 85.27] & \textbf{71.15 [71.01, 71.34]}   \\ 
\bottomrule
\end{tabular}
\label{tab:result}
\end{table*}

From the experimental results, we found that deep learning models do not always have an advantage over traditional manual scale ratings in the evaluation metrics.
Of the nine models we tested, only AttentionLSTM and our proposed method outperformed the manual approach, while the Transformer achieved comparable performance. This indicates that while manual scales are reliable, deep learning methods hold promising potential for future applications. Our proposed model is 2.4\% points higher in F1-score than the manual scale rating approach. This highlights the potential of leveraging advanced computational techniques in sensitive predictive tasks. 

When comparing the deep learning models internally, attention-based models like AttentionLSTM exhibit superior accuracy and precision, effectively prioritizing relevant features for more accurate predictions. Conversely, the GRU, BiGRU, Mamba, and our proposed model exhibit high recall, achieving greater than 85\%, which is approximately 7\% points higher than the traditional manual scale rating method. This is crucial for tasks where missing a true positive can have severe consequences. However, this higher recall comes with a trade-off of lower precision and potentially higher false positive rates.
Our proposed model achieves an optimal balance with the highest F1-score (71.15\%) among all tested models and competitive precision and recall. This balance is crucial for practical deployment in sensitive prediction tasks such as suicide prevention, suggesting that our model could serve as a highly effective tool for early detection and timely intervention. The results strongly advocate for the adoption of deep learning models in enhancing prediction capabilities beyond traditional methods.

\begin{table*}[!htb]
\centering
\caption{The top ten speech segments with the highest model attention weights in cases 1-4.}
\begin{tabular}{|l|l|} 
\hline
\textbf{Case (A)} & \begin{tabular}[c]{@{}l@{}}1. {Caller}: Ingested an excessive amount of sleeping pills due to a breakup with girlfriend.\\2. {Caller}: Daily depressive moods, particularly at night.\\3. {Caller}: Apathy towards all activities.\\4. {Caller}: History of self-harm with suicidal thoughts.\\5. {Caller}: One instance of attempted suicide from a balcony, intervened by father.\\6. {Operator}: Have you considered any suicide plans in the past week or two?\\7. {Caller}: Contemplating less painful methods of suicide.\\8. {Caller}: Forced self-feeding for over a week.\\9. {Caller}: Sleep reduced by approximately 5 hours in the past week.\\10. {Caller}: Silence, punctuated by sobs and sighs.\end{tabular}  \\ 
\hline
\textbf{Case (B)} & \begin{tabular}[c]{@{}l@{}}1. {Caller}: Persistent depressive mood over the past two to three months.\\2. {Caller}: Lost interest in life during the same period.\\3. {Caller}: Recurrent suicidal ideation.\\4. {Caller}: One act of attempted suicide.\\5. {Caller}: Currently sleeping only two hours per night, a decrease of three hours.\\6. {Caller}: Irritability sustained for over a week.\\7. {Caller}: Feeling worthless for two consecutive weeks.\\8. {Caller}: Attention deficit for four to five days and reduced libido for two weeks.\\9. {Caller}: Intense suffering.\\10. {Caller}: Frequent alcohol consumption without intoxication.\end{tabular}                                                       \\ 
\hline
\textbf{Case (C)} & \begin{tabular}[c]{@{}l@{}}1. {Caller}: Disillusioned with the world and parents.\\2. {Caller}: Consumes only one meal a day.\\3. {Caller}: Decline in perceived health.\\4. {Caller}: Five to six years of feeling worthless.\\5. {Caller}: Decreased libido in the past one to two years.\\6. {Caller}: Depression severely affecting personal and professional life.\\7. {Caller}: Engages in self-harm.\\8. {Caller}: Frequent worry about becoming aggressive towards others.\\9. {Operator}: Has someone attempted suicide and has it affected you?\\10. {Caller}: Mother's suicidal behavior impacting self-perception.\end{tabular}                                                                                           \\ 
\hline
\textbf{Case (D)} & \begin{tabular}[c]{@{}l@{}}1. {Caller}: Consistently low mood.\\2. {Caller}: Fatigued with life and harbors suicidal plans.\\3. {Caller}: History of three suicide attempts.\\4. {Caller}: Decreased appetite.\\5. {Operator}: Over the past two weeks, have there been any changes in your sleep?\\6. {Operator}: How much has your sleep duration increased compared to normal?\\7. {Caller}: Sleep has increased by 3 to 5 hours in the last two weeks.\\8. {Caller}: Minimal social interaction.\\9. {Caller}: Feeling lethargic for five months.\\10. {Caller}: Seven months of perceived worthlessness.\end{tabular}                                                                                                    \\
\hline
\end{tabular}
\label{tab:example}
\end{table*}

\section{Discussion} \label{sec:discussion}
% [Requirements]
% --------------------------------
% Dos  
% • 1st paragraph: Summarise and interpret the results in a simple manner. 
% • 2nd paragraph: Comparison with the existing literature, background, and any useful comments. 
% • 3rd paragraph: Outline biases and limitations, explaining in detail what the limitations are and how these can be addressed in the potential future studies. 
% • 4th paragraph: Short and straightforward conclusion, including a statement on the clinical implications/relevance of your study. Support your conclusion with the main results of your study. 

% Don’ts  
% • Don’t repeat information that is already stated in the Introduction. 
% • Don’t start the discussion with general considerations about the question/disease. 
% • Don’t introduce any new results. Discussion should be based only on the results stated in the previous chapter. 
% • Don’t make general statements or conclusions. 
% • Don’t make a hypothetical conclusion based on authors’ opinions and wishes.
% --------------------------------

Our study is based on the largest suicide hotline system in China, where we proposed a deep learning model for multi-task learning to predict future suicides. To the best of our knowledge, this is the first study to apply deep learning techniques to analyze psychological hotlines in the Chinese context. The proposed model achieves an F1-score of 71.15\%, outperforming traditional manual scale approaches and demonstrating significant potential for practical application. Unlike other research that relies on audio data and indirect measures of suicide risk~\cite{scherer2013investigating, amiriparian2024enhancing, belouali2021acoustic}, our study draws data directly from clinical settings, with suicide outcomes confirmed through long-term follow-up. 
Note that our study includes a significantly larger sample size, with 1,549 subjects compared to a maximum of 124 in other research.

As shown in Table~\ref{tab:example}, our findings reveal that the model not only focuses on data directly related to suicide but also extends its attention to include aspects such as patients' emotions, sleep patterns, and dietary habits. Specifically, the model concentrates on elements like suicide plans, methods, ideation, and expressions of existential distress. For instance, the first sentence of Case~(A) indicates that the caller has previously attempted suicide by consuming sleeping pills. 
Similarly, the second sentence of Case~(D) explicitly reveals that the caller has a formulated suicide plan. 
Additionally, Case~(A)~5,~7 and the Case~(B)~3 both underscore the caller's strong suicidal intent.
The model also captures expressions of intense distress. For instance Case~(B)~2,~9 and the Case~(D)~1 vividly convey the caller's intense expressions of distress. Notably, although the ninth sentence of Case~(A) does not explicitly articulate the caller's distress in words, the sobbing and sighing effectively communicate the depth of the caller's inner pain. According to Huang et al.~\cite{huang2019web} research, these factors are crucial for grading suicide risk.
Conversely, the model examines whether the caller has a history of suicidal behavior, as indicated in Case~(A)~5, the Case~(B)~4, and Case~(D)~3. According to the findings by Tong et al.~\cite{tong2024case}, the rate of subsequent suicide attempts among individuals with a history of attempted suicide is 100 times higher than that of the general population. Consequently, past instances of suicidal behavior serve as a significant indicator for predicting future suicide occurrences.
Lastly, the model also considers factors such as the caller's emotional state, sleep patterns, decreased libido, self-harm, and mental status, all of which are crucial in suicide risk assessment scales~\cite{tong2020prospective, tong2023predictive}. For instance, reduced sleep and forced self-feeding in Case~(A), persistent depressive mood and irritability in Case~(B), and disillusionment with the world and self-harm in Case~(C), all provide vital context for assessing the caller's mental health and suicide risk.

%However, our research still have some limitations. From the model construction standpoint, we segmented the speech every thirty seconds, which may disrupt the complete dialogue and result in incomplete information. Although the attention mechanism in the Transformer we used may help build coherent information, the exact principle mechanism remains to be investigated. Additionally, while we conducted follow-up psychosocial support, our inability to know with certainty the exact time of suicide for callers who committed suicide limits us to construct fine-grained models. 
% However, from another perspective, if our model can quickly identify high-risk individuals, it may be sufficient to focus more frequent follow-ups on these high-risk targets, making the need for precise suicide prediction less critical.
However, our study has some limitations. Firstly, the ratio of positive and negative cases in the dataset does not reflect the actual proportions. In real life, approximately 8.4\% of individuals commit suicide, whereas in our experimental setting, this proportion is 48.16\%. Of course, this does not imply that our study is biased, as the purpose of our research is model development, with validation performed on a separate test set. The next step is to validate the performance of our model in real-world prospective experiments.
% However, compare with other study like Scherer et al.~\cite{scherer2013investigating} and Amiriparian et al.~\cite{amiriparian2024enhancing} which were use leave-one-subject-out validation strategy, our study is much more better.
Also, the models we developed were focused on suicide prediction with a twelve-month follow-up, without considering specific periods, such as which months the risk is highest. In our next step, we will aim to construct a fine-grained model that accounts for these specific time frames. 
% Finally, this study remains at the stage of validating existing datasets and has not conducted prospective research to assess the extent to which these evaluation techniques can accurately predict suicide risk.
%As a next step, we plan to use trained models to analyze the characteristics of different populations (e.g., different age groups) to better understand their psychological characteristics and principles. Additionally, we need to focus on researching whether the model is biased across different populations, such as different genders and dialects, to ensure the development of an unbiased model. Currently, our model primarily processes and analyzes voice data, and we aim to use a multimodal approach that converts audio to text for joint modeling, thereby improving the accuracy of suicide prediction in the future. 
The proposed model achieved higher performance than the manual scale rating approach. However, we are also interested in conducting research on human-machine collaboration by integrating deep learning suicide prediction results with manual scale assessments to ultimately enhance the prediction of suicide risk. 

\section{Conclusion} \label{sec:conclusion}
This research explores the application of deep learning technology in predicting suicide risk using long-sequence speech data from psychological support hotlines.
We conducted experiments with a large population and proposed new models to address long speech, aiming to identify the impact of detailed emotions and emotional changes on future suicides. 
The experimental results demonstrate that our proposed model achieved a higher F1-score, with an improvement of 2.4\% points over the traditional manual scale rating method, and it outperformed the eight compared deep learning models.
This finding not only underscores the potential of deep learning in this field but also sets a new performance benchmark for future research.
Our future work will focus on prospectively validating the performance of our method in a clinical setting and investigating whether our proposed method can improve the efficiency of risk identification in psychological support hotlines.

\section{Acknowledgments}\label{sec:acknowledgments}
This study was supported by the National Natural Science Foundation of China [82071546], Beijing Municipal High Rank Public Health Researcher Training Program [2022-2-027], the Beijing Hospitals Authority Clinical Medicine Development of Special Funding Support [ZYLX202130], and the Beijing Hospitals Authority’s Ascent Plan [DFL20221701]. 
The funding institutions had no role in the design, data collection, analysis, interpretation, writing of the report, or the decision to submit for publication.
Guanghui Fu is supported by a Chinese Government Scholarship provided by the China Scholarship Council (CSC).

\section{Thical Statement}
We value participant privacy and are committed to protecting the security of their personal information, ensuring anonymity and confidentiality throughout the data processing phase. 
All analyses in this study are aimed at enhancing scientific knowledge and social welfare, with no commercial or financial conflicts of interest. 
The research team takes full responsibility for all content and conclusions presented in this study.
Before callers were connected to a hotline operator, they were informed by a voice message that all calls of the hotline would be tape-recorded and
that data utilized for research purposes would be analyzed anonymously.
Caller’s personal information were noted for clinical purposes but deleted prior to data analysis. All callers assessed as high-risk (or operator considered necessary for other callers) received a semistructured
hotline-based intervention during the index call.
\bibliographystyle{IEEEtran}
\bibliography{ref}

%Biography text here without a photo.

%\begin{IEEEbiography}[{\includegraphics[width=1in,height=1.25in,clip,keepaspectratio]{figs/songchangwei.jpg}}]{IEEE Publications Technology Team}
%In this paragraph you can place your educational, professional background and research and other interests.\end{IEEEbiography}

\end{document}